\documentclass[10pt,journal,cspaper,compsoc]{IEEEtran}
\pdfoutput = 1
\usepackage{epsfig}
\usepackage{epstopdf}
\usepackage{graphicx}
\usepackage{amsmath}
\usepackage{amssymb}
\usepackage{subfigure}
\usepackage{indentfirst} 

\newcommand{\etal}{\textit{et al.}}
\newtheorem{proposition}{Proposition}

\ifCLASSOPTIONcompsoc
\else
\fi
\ifCLASSINFOpdf
\else
\fi

\begin{document}
	%
	\title{Reconstruction-free action inference from compressive imagers}
	%
	%
	%
	%
	
	\author{Kuldeep~Kulkarni,
		Pavan~Turaga
		\IEEEcompsocitemizethanks{\IEEEcompsocthanksitem K. Kulkarni and P. Turaga are with the School of Arts, Media and Engineering and School of Electrical, Computer and Energy Engineering, Arizona State University. Email: kkulkar1@asu.edu, pturaga@asu.edu.\protect\\
		}
		\thanks{}}
	
	%
	%


	\IEEEcompsoctitleabstractindextext{%
		\begin{abstract}
			Persistent surveillance from camera networks, such as at parking lots, UAVs, etc., often results in large amounts of video data, resulting in significant challenges for inference in terms of storage, communication and computation. Compressive cameras have emerged as a potential solution to deal with the data deluge issues in such applications. However, inference tasks such as action recognition require high quality features which implies reconstructing the original video data. Much work in compressive sensing (CS) theory is geared towards solving the reconstruction problem, where state-of-the-art methods are computationally intensive and provide low-quality results at high compression rates. Thus, reconstruction-free methods for inference are much desired. In this paper, we propose reconstruction-free methods for action recognition from compressive cameras at high compression ratios of 100 and above.  Recognizing actions directly from CS measurements requires features which are mostly nonlinear and thus not easily applicable. This leads us to search for such properties that are preserved in compressive measurements. To this end, we propose the use of spatio-temporal smashed filters, which are compressive domain versions of pixel-domain matched filters. We conduct experiments on publicly available databases  and show that one can obtain recognition rates that are comparable to the oracle method in uncompressed setup, even for high compression ratios.
		\end{abstract}
		
		}

		\maketitle

		\IEEEdisplaynotcompsoctitleabstractindextext
		
		\IEEEpeerreviewmaketitle

		\section{Introduction}
		Action recognition is one of the long standing research areas in computer vision with widespread applications in video surveillance, unmanned aerial vehicles (UAVs), and real-time monitoring of patients. All these applications are heavily resource-constrained and require low communication overheads in order to achieve real-time implementation. Consider the application of UAVs which provide real-time video and high resolution aerial images on demand. In these scenarios, it is typical to collect an enormous amount of data, followed by transmission of the same to a ground station using a low-bandwidth communication link. This results in expensive methods being employed for video capture, compression, and transmission implemented on the aircraft. The transmitted video is decompressed at a central station and then fed into a action recognition pipeline. Similarly, a video surveillance system which typically employs many high-definition cameras, gives rise to a prohibitively large amount of data, making it very challenging to store, transmit and extract meaningful information. Thus, there is a growing need to acquire as little data as possible and yet be able to perform high-level inference tasks like action recognition reliably.
		\par
		
		Recent advances in the areas of compressive sensing (CS) \cite{Candes} have led to the development of new sensors like compressive cameras (also called single-pixel cameras (SPCs)) \cite{SPC} greatly reduce the amount of sensed data, yet preserve most of its information. More recently, InView Technology Corporation applied CS theory to build commercially available CS workstations and SWIR (Short Wave Infrared) cameras, thus equipping CS researchers with a hitherto unavailable armoury to conduct experiments on real CS imagery. In this paper, we wish to investigate the utility of compressive cameras for action recognition in improving the tradeoffs between reliability of recognition and computational/storage load of the system in a resource constrained setting. CS theory states that if a signal can be represented by very few number of coefficients in a basis, called the sparsifying basis, then the signal can be reconstructed nearly perfectly even in the presence of noise, by sensing sub-Nyquist number of samples \cite{Candes}. SPCs differ from conventional cameras in that 
		\begin{figure}[ht!]
			\centering
			\includegraphics[width=0.4\textwidth, height=18.5em]{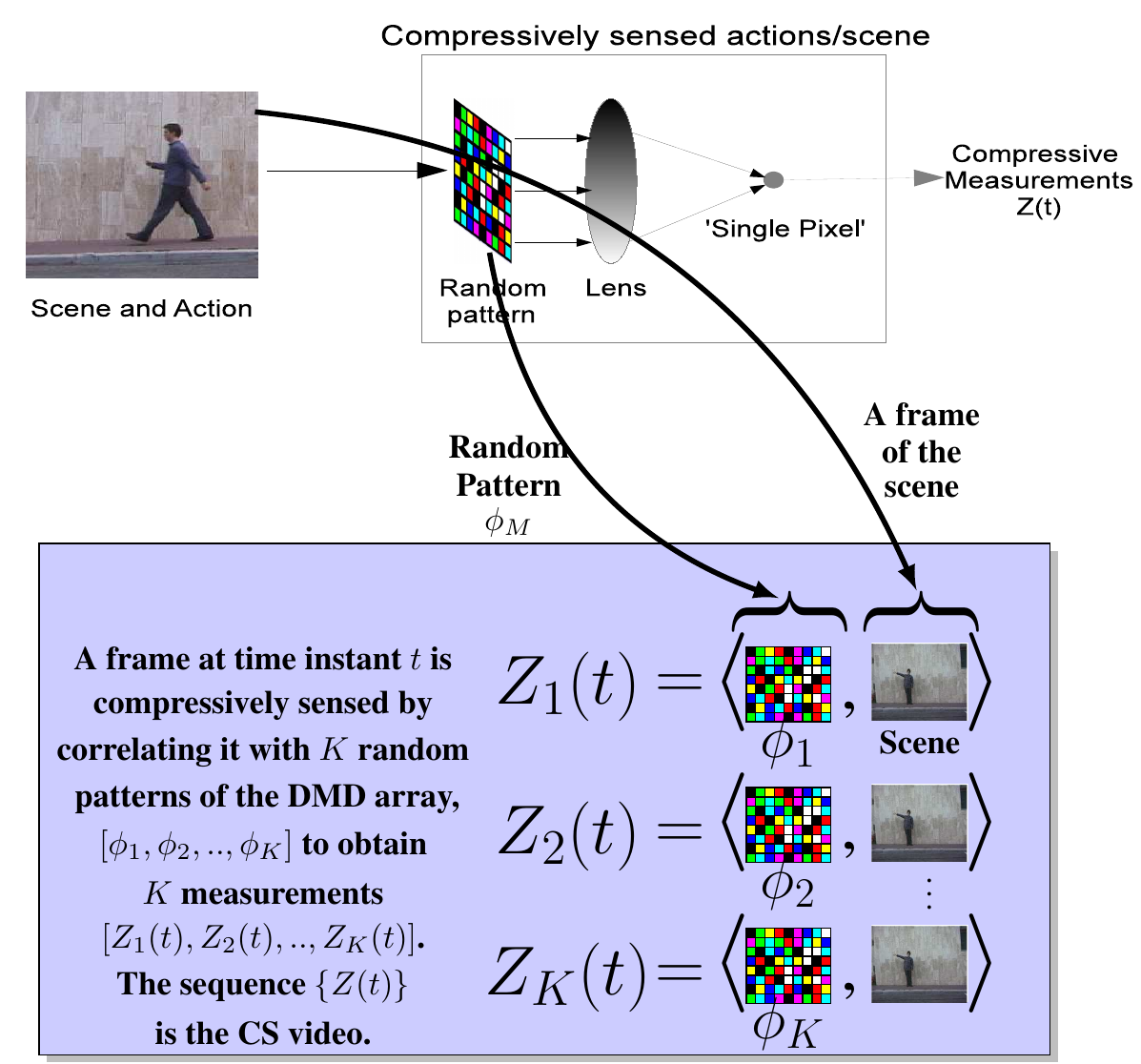}
			\caption{\small{Compressive Sensing (CS) of a scene: Every frame of the scene is compressively sensed by optically correlating random patterns with the frame to obtain CS measurements. The temporal sequence of such CS measurements is the CS video.}}
			\label{fig:CSvideo}
			\vspace{-0.3cm}
		\end{figure}
		they integrate the process of acquisition and compression by acquiring a small number of linear projections of the original images. More formally, when a sequence of images is acquired by a compressive camera, the measurements are generated by a sensing strategy which maps the space of $P \times Q$ images, $I \in \mathbb{R}^{PQ}$ to an observation space $Z \in \mathbb{R}^{K}$, 
		
		\begin{equation}
		Z(t) = \phi I(t) + w(t), \label{eq:formulation}
		\end{equation} 
		where $\phi$ is a $K \times PQ$ measurement matrix, $w(t)$ is the noise, and $K \ll PQ$. The process is pictorially shown in Figure \ref{fig:CSvideo}.
		
		\paragraph*{{\bf Difference between CS and video codecs}} It is worth noting at this point that the manner in which compression is achieved by SPCs differs fundamentally from the manner in which compression is achieved in JPEG images or MPEG videos. In the case of JPEG, the images are fully sensed and then compressed by applying wavelet transform or DCT to the sensed data, and in the case of MPEG, a video after having been sensed fully is compressed using a motion compensation technique. However, in the case of SPCs, at the outset one does not have direct access to full blown images, $\{I(t)\}$. SPCs instead provide us with compressed measurements $\{Z(t)\}$ directly by optically calculating inner products of the images, $\{I(t)\}$, with a set of test functions given by the rows of the measurement matrix, $\phi$, implemented using a programmable micro-mirror array \cite{SPC}.  While this helps avoid the storage of a large amount of data and expensive computations for compression, it often comes at the expense of employing high computational load at the central station to reconstruct the video data perfectly. Moreover, for perfect reconstruction of the images, given a sparsity level of $s$, state-of-the-art algorithms require $O(s\log(PQ/s))$ measurements \cite{Candes}, which still amounts to a large fraction of the original data dimensionality. Hence, using SPCs may not always provide advantage with respect to communication resources since compressive measurements and transform coding of data require comparable bandwidth~\cite{Dikpal}. However, we show that it is indeed possible to perform action recognition at much higher compression ratios, by bypassing reconstruction. In order to do this, we propose a spatio-temporal smashed filtering approach, which results in robust performance at extremely high compression ratios.

		\vspace{-0.2cm}
		\subsection{Related work}
		\par
		\paragraph*{\textbf{a) Action Recognition}}
		The approaches in human action recognition from cameras can be categorized based on the low level features. Most successful representations of human action are based on features like optical flow, point trajectories, background subtracted blobs and shape, filter responses, etc. Mori \etal \cite{Mori1} and Cheung \etal \cite{Cheung} used geometric model based and shape based representations to recognize actions. Bobick and Davis \cite{bobick2001recognition} represented actions using 2D motion energy and motion history images from a sequence of human silhouettes. Laptev \cite{Laptev} extracted local interest points from a 3-dimensional spatiotemporal volume, leading to a concise representation of a video. Wang \etal \cite{wang2009evaluation} evaluated various combinations of space-time interest point detectors (Harris3D, Cuboid, Hessian) and several descriptors to perform action recognition. The current state-of-the-art approaches \cite{wang2011action,wang2013action} to action recognition are based on dense trajectories, which are extracted using dense optical flow. The dense trajectories are encoded by complex, hand-crafted descriptors like histogram of oriented gradients (HOG) \cite{dalal2005histograms} , histogram of oriented optical flow (HOOF) \cite{chaudhry2009histograms}, HOG3D \cite{klaser2008spatio}, and motion boundary histograms (MBH) \cite{wang2011action}. For a detailed survey of action recognition, the readers are referred to \cite{aggarwal2011human}. However, the extraction of the above features involves various non-linear operations. This makes it very difficult to extract such features from compressively sensed images.
		\par 
		
		\paragraph*{\textbf{b) Action recognition in compressed domain}}
		Though action recognition has a long history in computer vision, little exists in literature to recognize actions in the compressed domain. Yeo \etal \cite{Kannan} and Ozer \etal \cite{Ozer:2000:HAD:822088.823430} explore compressed domain action recognition from MPEG videos by exploiting the structure, induced by the motion compensation technique used for compression. MPEG compression is done on a block-level which preserves some local properties in the original measurements. However, as stated above, the compression in CS cameras is achieved by randomly projecting the individual frames of the video onto a much lower dimensional space and hence does not easily allow leveraging motion information of the video. CS imagery acquires global measurements, thereby do not preserve any local information in their raw form, making action recognition much more difficult in comparison.

		\paragraph*{\textbf{c) Inference problems from CS video}} Sankaranarayanan \etal \cite{CSLDS} attempted to model videos as a LDS (Linear Dynamical System) by recovering parameters directly from compressed measurements, but is sensitive to spatial and view transforms, making it more suitable for recognition of dynamic textures than action recognition. Thirumalai \etal \cite{EPFL} introduced a reconstruction-free framework to obtain optical flow based on  correlation estimation between two compressively sensed images. However, the method does not work well at very low measurement rates. Calderbank \etal \cite{CLearning} theoretically showed that `learning directly in compressed domain is possible', and that with high probability the linear kernel SVM classifier in the compressed domain can be as accurate as best linear threshold classifier in the data domain. Recently, Kulkarni and Turaga \cite{Rectex} proposed a novel method based on recurrence textures for action recognition from compressive cameras. However, the method is prone to produce very similar recurrence textures even for dissimilar actions for CS sequences and is more suited for feature sequences as in \cite{SelfSimilar}. 
		\begin{figure*}[ht!]
			\centering
			\hspace{-2.2cm}
			\includegraphics[width=0.9\textwidth,height=27em]{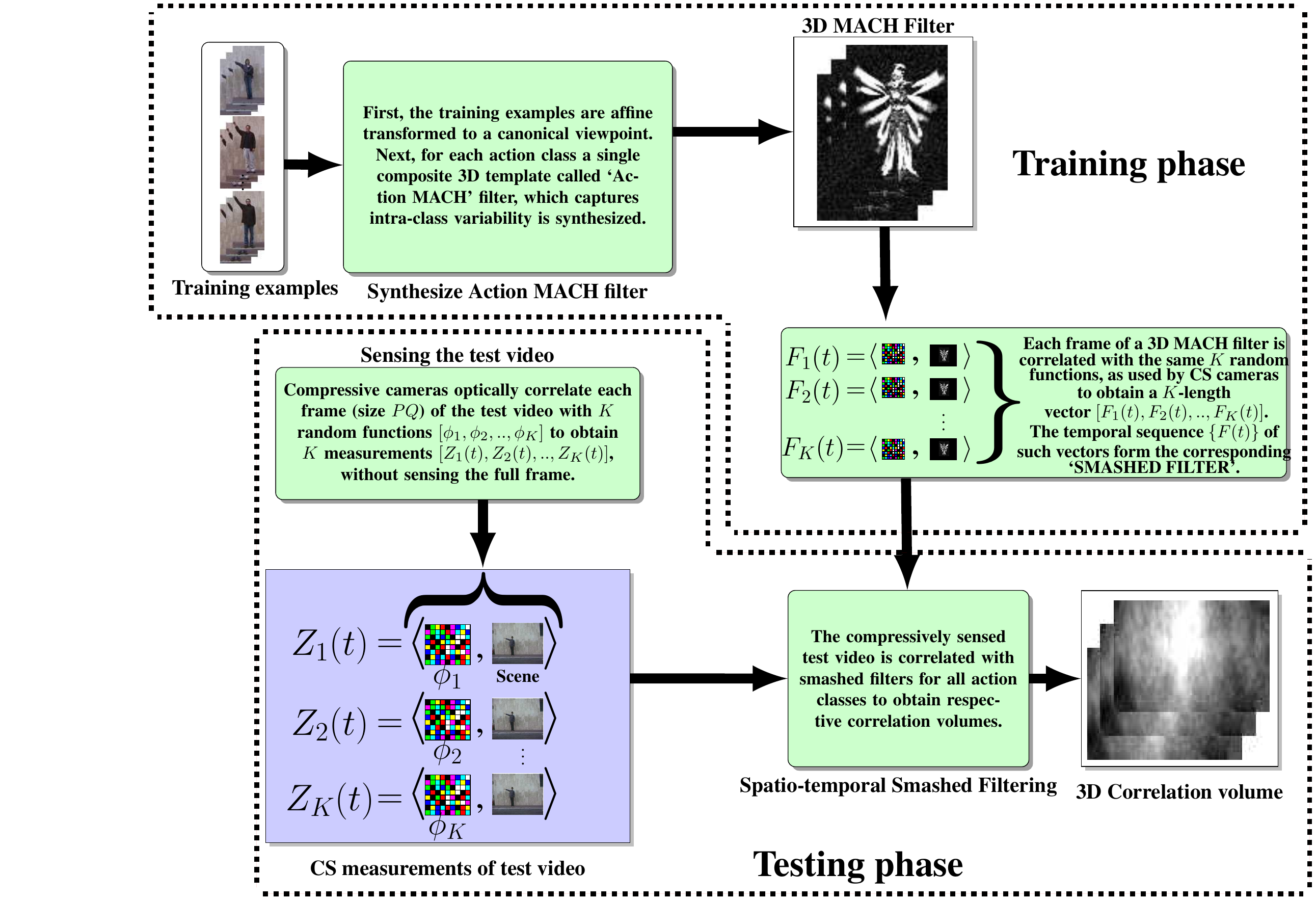}
			\caption{\small{Overview of our approach to action recognition from a compressively sensed test video. First, MACH \cite{MACH} filters for different actions are synthesized offline from training examples and then compressed to obtain smashed filters. Next, the CS measurements of the test video are correlated with these smashed filters to obtain correlation volumes which are analyzed to determine the action in the test video.}}
			\label{fig:Overview}
			\vspace{-0.4cm}
		\end{figure*}

		\paragraph*{\textbf{d) Correlation filters in computer vision}}
		Even though, as stated above, the approaches based on dense trajectories extracted using optical flow information have yielded state-of-the-art results, it is difficult to extend such approaches while dealing with compressed measurements. Earlier approaches to action recognition were based on correlation filters, which were obtained directly from pixel data  \cite{kim2009canonical,shechtman2005space,seo2011action,MACH,YorkAc2010,AcBank2012}. The filters for different actions are correlated with the test video and the responses thus obtained are analyzed to recognize and locate the action in the test video. Davenport \etal \cite{Davenport} proposed a CS counterpart of the correlation filter based framework for target classification. Here, the trained filters are compressed first to obtain `smashed filters', then the compressed measurements of the test examples are correlated with these smashed filters. Concisely, smashed filtering hinges on the fact that correlation between a reference signal and an input signal is nearly preserved even when they are projected onto a much lower-dimensional space. 
		In this paper, we show that spatio-temporal smashed filters provide a natural solution to reconstruction-free action recognition from compressive cameras. Our framework (shown in Figure \ref{fig:Overview}) for classification includes synthesizing Action MACH (Maximum Average Correlation Height) filters \cite{MACH} offline and then correlating the compressed versions of the filters with compressed measurements of the test video, instead of correlating raw filters with full-blown video, as is the case in \cite{MACH}. Action MACH involves synthesizing a single 3D spatiotemporal filter which captures information about a specific action from a set of training examples. MACH filters can become ineffective if there are viewpoint variations in the training examples. To effectively deal with this problem, we also propose a quasi view-invariant solution, which can be used even in uncompressed setup.

		\textbf{Contributions:} 1) We propose a correlation-based framework for action recognition and localization directly from compressed measurements, thus avoiding the costly reconstruction process.  2) We provide principled ways to achieve quasi view-invariance in a spatio-temporal smashed filtering based action recognition setup. 3) We further show that a single MACH filter for a canonical view is sufficient to generate MACH filters for all affine transformed views of the canonical view. 
		\par
		\textbf{Outline:} Section 2 outlines the reconstruction-free framework for action recognition, using spatio-temporal smashed filters (STSF). In section 3, we describe a quasi view-invariant solution to MACH based action recognition by outlining a simple method to generate MACH filters for any affine transformed view. In section 4, we present experimental results obtained on three popular action databases, Weizmann, UCF sports, UCF50 and HMDB51 databases.

		\section{Reconstruction free action recognition}
		
		To devise a reconstruction-free method for action recognition from compressive cameras, we need to exploit such properties that are preserved robustly even in the compressed domain. One such property is the distance preserving property of the measurement matrix $\phi$ used for compressive sensing \cite{Candes,JL}. Stated differently, the correlation between any two signals is nearly preserved even when the data is compressed to a much lower dimensional space. This makes correlation filters a natural choice to adopt. 2D correlation filters have been widely used in the areas of automatic target recognition and biometric applications like face recognition \cite{FACE}, palm print identification \cite{PALM}, etc., due to their ability to capture intraclass variabilities. Recently, Rodriguez \etal \cite{MACH} extended this concept to 3D by using a class of correlation filters called MACH filters to recognize actions. As stated earlier, Davenport \etal \cite{Davenport} introduced the concept of smashed filters by implementing matched filters in the compressed domain. In the following section, we generalize this concept of smashed filtering to the space-time domain and show how 3D correlation filters can be implemented in the compressed domain for action recognition.

		\subsection{Spatio-temporal smashed filtering (STSF)}\label{STSF}
		
		This section forms the core of our action recognition pipeline, wherein we outline a general method to implement spatio-temporal correlation filters using compressed measurements without reconstruction and subsequently, recognize actions using the response volumes. To this end, consider a given video $s(x,y,t)$ of size $P \times Q \times R$ and let $H_{i}(x,y,t)$ be the optimal 3D matched filter for actions $i = 1,..,N_A$, with size $L \times M \times N$ and $N_A$ is the number of actions. First, the test video is correlated with the matched filters of all actions $i = 1,..N_A$ to obtain respective 3D response volumes as in \eqref{response}.
		\vspace{-0.2cm}
		\begin{equation}
		c_i(l,m,n) = \sum_{t=0}^{N-1} \sum_{y=0}^{M-1} \sum_{x=0}^{L-1} s(l+x,m+y,n+t)H_i(x,y,t).
		\label{response}
		\end{equation} 
		Next, zero-padding each frame in $H_i$ upto a size $P \times Q$ and changing the indices, \eqref{response} can be rewritten as:
		\vspace{-0.1cm}
		\begin{equation}\label{response1}
		c_i(l,m,n) = \sum_{t=0}^{N-1} \sum_{\beta=0}^{Q-1} \sum_{\alpha=0}^{P-1}s(\alpha,\beta,n+t)H_i(\alpha-l,\beta-m,t).
		\end{equation}
		
		This can be written as the summation of $N$ correlations in the spatial domain as follows: 
		\vspace{-0.1cm}
		\begin{equation}\label{eq:corr}
		\vspace{-0.3cm}
		c_i(l,m,n) = \sum_{t=0}^{N-1} \langle S_{n+t} , H_i^{l,m,t} \rangle,
		\end{equation}  where, $\langle , \rangle$ denotes the dot product, $S_{n+t}$ is the column vector obtained by concatenating the $Q$ columns of the $(n+t)^{th}$ frame of the test video. To obtain $H_i^{l,m,t}$, we first shift the $t^{th}$ frame of the zeropadded filter volume $H_i$ by $l$ and $m$ units in $x$ and $y$ respectively to obtain an intermediate frame and then rearrange it to a column vector by concatenating its $Q$ columns. Due to the distance preserving property of measurement matrix $\phi$, the correlations are nearly preserved in the much lower dimensional compressed domain. To state the property more specifically, using JL Lemma \cite{JL}, the following relation can be shown:
		\vspace{-0.2cm}
		\begin{equation}\label{eq:error1}
		c_i(l,m,n) - N\epsilon \le  \sum_{t=0}^{N-1} \langle \phi S_{n+t} , \phi H_i^{l,m,t} \rangle \le c_i(l,m,n) + N\epsilon  .
		\end{equation}
		The  derivation of this relation and the precise form of $\epsilon$ is as follows. In the following, we derive the relation between the response volume from uncompressed data and response volume obtained using compressed data.
		According to JL Lemma \cite{JL}, given $0< \epsilon < 1$ , a set $\mathcal{S}$ of $2N$ points in $\mathbb{R}^{PQ}$, each with unit norm, and $K > \mathcal{O}(\frac{log(N)}{\epsilon ^{2}})$, there exists a Lipschitz function $f: \mathbb{R}^{PQ} \rightarrow \mathbb{R}^{K}$ such that 
		\vspace{-0.2cm}
		\begin{eqnarray}\label{eq:error2}
		(1-\epsilon)\|S_{n+t}-H_i^{l,m,t}\|^{2} &\le \|f(S_{n+t})- f(H_i^{l,m,t})\|^{2} \nonumber \\ & \le   (1+\epsilon)\|S_{n+t}-H_i^{l,m,t}\|^{2} \nonumber , \\ &
		\end{eqnarray}
		and
		\begin{eqnarray}\label{eq:error3}
		(1-\epsilon)\|S_{n+t}+H_i^{l,m,t}\|^{2} &\le \|f(S_{n+t})+ f(H_i^{l,m,t})\|^{2} \nonumber \\ & \le  (1+\epsilon)\|S_{n+t}+H_i^{l,m,t}\|^{2} \nonumber , \\ &
		\end{eqnarray}
		
		$\forall$ $S_{n+t}$ and $H_i^{l,m,t}$  $\in \mathcal{S}$. 
		Now we have:
		\vspace{-0.2cm}
		\begin{eqnarray}
		4 \langle f(S_{n+t}) , f(H_i^{l,m,t})\rangle  \nonumber \\ = \|f(S_{n+t})+ f(H_i^{l,m,t})\|^{2} - \|f(S_{n+t})- f(H_i^{l,m,t})\|^{2}  \nonumber  \\ 
		\ge (1+\epsilon)\|S_{n+t}+H_i^{l,m,t}\|^{2} - (1+\epsilon)\|S_{n+t}-H_i^{l,m,t}\|^{2} \nonumber  \\
		= 4 \langle S_{n+t} ,  H_i^{l,m,t} \rangle - 2\epsilon(\|S_{n+t}\|^{2}+\|H_i^{l,m,t}\|^{2}) \nonumber  \\
		\ge 4 \langle S_{n+t} ,  H_i^{l,m,t}\rangle - 4\epsilon.  \nonumber \\ &
		\label{eq:error4}
		\end{eqnarray}
		We can get a similar relation for opposite direction, which when combined with \eqref{eq:error4}, yields the following:
		\begin{eqnarray}\label{eq:error5}
		\langle S_{n+t} , H_i^{l,m,t} \rangle - \epsilon  \le \langle f(S_{n+t}) , f(H_i^{l,m,t}) \rangle 
		\nonumber \\  \le \langle S_{n+t} , H_i^{l,m,t} \rangle + \epsilon.
		\end{eqnarray}
		However, JL Lemma does not provide us with a embedding, $f$ which satisfies the above relation. As discussed in \cite{Achlioptas2001}, $f$ can be constructed as a matrix, $\phi$ with size $K \times PQ$, whose entries are either
		\begin{itemize}
			\item independent realizations of Gaussian random variables or
			\item independent realizations of $\pm$ Bernoulli random variables.
		\end{itemize}
		Now, if $\phi$ constructed as explained above is used as measurement matrix, then we can replace $f$ in \eqref{eq:error5} by $\phi$, leading us to
		\begin{eqnarray}\label{eq:error6}
		\langle S_{n+t} , H_i^{l,m,t} \rangle - \epsilon  \le \langle \phi S_{n+t} , \phi H_i^{l,m,t} \rangle \nonumber \\ \le \langle S_{n+t} , H_i^{l,m,t} \rangle + \epsilon .
		\end{eqnarray}
		Hence, we have,
		\vspace{-0.2cm}
		\begin{eqnarray}
		\sum_{t=0}^{N-1} \langle S_{n+t} , H_i^{l,m,t} \rangle - N\epsilon  \le \sum_{t=0}^{N-1} \langle \phi S_{n+t} , \phi H_i^{l,m,t} \rangle  \nonumber
		\vspace{-8mm}
		\end{eqnarray}
		\begin{equation}\label{eq:error6}
		\le \sum_{t=0}^{N-1} \langle S_{n+t} , H_i^{l,m,t} \rangle + N\epsilon.
		\end{equation}
		Using equations (4) and \eqref{eq:error6}, we arrive at the following desired equation.
		\begin{equation}\label{eq:error7}
		c_i(l,m,n) - N\epsilon \le  \sum_{t=0}^{N-1} \langle \phi S_{n+t} , \phi H_i^{l,m,t} \rangle  \le c_i(l,m,n) + N\epsilon  .
		\end{equation}
		Now allowing for the error in correlation, we can compute the response from compressed measurements as below:
		
		\begin{equation}
		\vspace{-0.25cm}
		c_i^{comp}(l,m,n) = \sum_{t=0}^{N-1} \langle \phi S_{n+t} , \phi H_i^{l,m,t} \rangle.
		\end{equation}
		
		The above relation provides us with the 3D response volume for the test video with respect to a particular action, without reconstructing the frames of the test video. To reduce computational complexity, the 3D response volume is calculated in frequency domain via 3D FFT. 
		\paragraph*{{\bf Feature vector and Classification using SVM}}
		For a given test video, we obtain $N_A$ correlation volumes. For each correlation volume, we adapt three level volumetric max-pooling to obtain a 73 dimensional feature vector \cite{AcBank2012}. In addition, we also compute peak-to-side-lobe-ratio for each of these 73 maxpooled
		values. PSR is given by $PSR_k = \frac{peak_{i}-\mu_{i}}{\sigma_{i}}$ ,where $peak_k$ is the $k^{th}$ max-pooled value, and $\mu_k$ and $\sigma_k$ are the mean and standard deviation values in its small neighbourhood. Thus, the feature vector for a given test video is of dimension, $N_A \times 146$. This framework can be used in any reconstruction-free
		application from compressive cameras which can be implemented using 3D correlation filtering. Here, we assume that there exists an optimal matched filter for each action and outline a way to recognize actions from compressive measurements. In the next section, we show how these optimal filters are obtained for each action.
		
		\subsection{Training filters for action recognition} 
		The theory of training correlation filters for any recognition task is based on synthesizing a single template from training examples, by finding an optimal tradeoff between certain performance measures. Based on the performance measures, there exist a number of classes of correlation filters.  A MACH filter is a single filter that encapsulates the information of all training examples belonging to a particular class and is obtained by optimizing four performance parameters, the Average Correlation Height (ACH), the Average Correlation Energy (ACE), the Average Similarity Measure (ASM), and the Output Noise Variance (ONV). Until recently, this was used only in two dimensional applications like palm print identification \cite{PALM}, target recognition \cite{target1} and face recognition problems \cite{FACE}. For action recognition, Rodriguez et al. \cite{MACH} introduced a generalized form of MACH filters to synthesize a single action template from the spatio-temporal volumes of the training examples. Furthermore, they extended the notion for vector-valued data. In our framework for compressive action recognition, we adopt this approach to train matched filters for each action. Here, we briefly give an overview of 3D MACH filters which was first described in \cite{MACH}. 
		\par
		First, temporal derivatives of each pixel in the spatio-temporal volume of each training sequence are computed and the frequency domain representation of each volume is obtained by computing a 3D-DFT of that volume, according to the following:
		
		\begin{equation}
		\label{eq:FT1}
		F({\bf u}) =  \sum_{t=0}^{N-1} \sum_{x_{2}=0}^{M-1} \sum_{x_{1}=0}^{L-1} f({\bf x}) e^{(-j2\pi({\bf u \cdot x}))} ,
		\vspace{-0.1cm}
		\end{equation} 
		where, $f({\bf x})$ is the spatio-temporal volume of $L$ rows, $M$ columns and $N $ frames, $F({\bf u})$ is its spatio-temporal representation in the frequency domain and  ${\bf x} = (x_{1},x_{2},t)$ and ${\bf u} = (u_1,u_2,u_3)$ denote the indices in space-time and frequency domain respectively. If $N_{e}$ is the number of training examples for a particular action, then we denote their 3D DFTs by $X_{i}({\bf u}), i = 1,2,.., N_{e}$, each of dimension, $d = L\times M \times N$. The average spatio-temporal volume of the training set in the frequency domain is given by
		$
		M_{x}({\bf u}) = \frac{1}{N_e} \sum_{i=1}^{N_e}X_{i}({\bf u}).
		$
		The average power spectral density of the training set is given by
		$
		D_x({\bf u}) = \frac{1}{N_e}\sum_{i=1}^{N_e} |X_{i}({\bf u})|^2,
		$ and the average similarity matrix of the training set is given by
		$
		S_x({\bf u}) = \frac{1}{N_e}\sum_{i=1}^{N_e} |X_{i}({\bf u})-M_{x}({\bf u})|^2.
		$ 
		Now, the MACH filter for that action is computed by minimizing the average correlation energy, average similarity measure, output noise variance and maximizing the average correlation height. This is done by computing the following:
		\begin{equation}\label{eq:filter}
		h({\bf u}) = \frac{1}{[\alpha C({\bf u}) + \beta D_x({\bf u}) + \gamma S_x({\bf u})]} M_x({\bf u}),
		\end{equation}
		where, $C({\bf u})$ is the noise variance at the corresponding frequency. Generally, it is set to be equal to 1 at all frequencies. The corresponding space-time domain representation $H(x,y,t)$ is obtained by taking the inverse 3D DFT of $h$. A filter with response volume $H$ and parameters $\alpha$, $\beta$ and $\gamma$ is compactly written as ${\bf H} = \{H,\alpha,\beta,\gamma\}$.
		
		\section{Affine Invariant Smashed Filtering} 
		
\begin{figure}[ht!]
			\centering
			\subfigure[]{\includegraphics[width=0.5\textwidth]{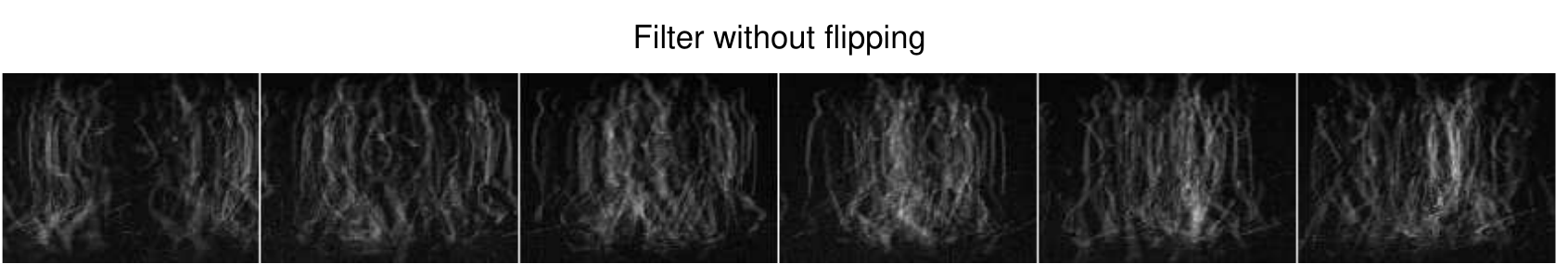}
				\label{fig:subfig1}
			}
			
			\subfigure[]{\includegraphics[width=0.5\textwidth]{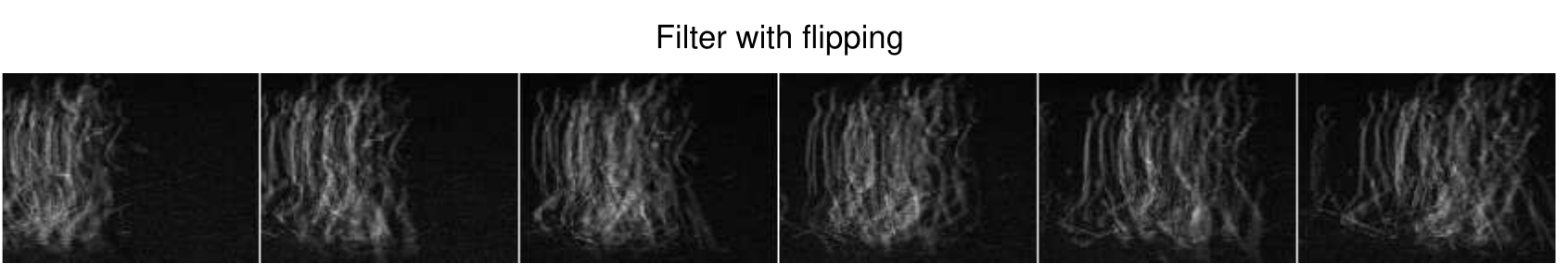}
				\label{fig:subfig2}
			}
			\caption{\small{a) `Type-1' filter obtained for walking action where the training examples were from different viewpoints b) `Type-2' filter obtained from the training examples by bringing all the training examples to the same viewpoint. In (a), two groups of human move in opposite directions and eventually merge into each other, thus making the filter ineffective. In (b), the merging effect is countered by transforming the training set to the same viewpoint.}}
			\label{fig:filtercomp}
			
		\end{figure}

		Even though MACH filters capture intra-class variations, the filters can become ineffective if viewpoints of the training examples are different or if the viewpoint of the test video is different from viewpoints of the training examples. Filters thus obtained may result in misleading correlation peaks. Consider the case of generating a filter of a translational action, walking, wherein the training set is sampled from two different views. The top row in Fig \ref{fig:filtercomp} depicts some frames of the filter, say `Type-1' filter, generated out of such a training set. The bottom row depicts some frames of the filter, say `Type-2' filter, generated by affine transforming all examples in the training set to a canonical viewpoint. 
		Roughly speaking, the `Type-2' filter can be interpreted as many humans walking in the same direction, whereas the `Type-1' filter, as 2 groups of humans, walking in opposite directions. One can notice that some of the frames in the `Type-1' do not represent the action of interest, particularly the ones in which the two groups merge into each other. This kind of merging effect will become more prominent as the number of different views in the training set increases. The problem is avoided in the `Type-2' filter because of the single direction of movement of the whole group. Thus, it can be said that the quality of information about the action in the `Type-2' filter is better than that in the `Type-1' filter. As we show in experiments, this is indeed the case. Assuming that all views of all training examples are affine transforms of a canonical view, we can synthesize a MACH filter generated after transforming all training examples to a common viewpoint and avoid the merging effect. However, different test videos may be in different viewpoints, which makes it impractical to synthesize filters for every viewpoint. Hence it is desirable that a single representative filter be generated for all affine transforms of a canonical view. The following proposition asserts that, from a MACH filter defined for the canonical view, it is possible to obtain a compensated MACH filter for any affine transformed view.
		
		\begin{proposition}
			
			Let ${\bf H} = \{H,\alpha,\beta,\gamma\}$ denote the MACH filter in the canonical view, then for any arbitrary view $V$, related to the canonical view by an affine transformation, $[A|{\bf b}]$, there exists a MACH filter, ${\bf \hat{H}} =\{\hat{H},\hat{\alpha},\hat{\beta},\hat{\gamma}\}$  such that: 
			$\hat{H}({\bf x_{s}},t) = {|\Delta|^2}H(A{\bf x_{s}}+{\bf b},t)$, $\hat{\alpha} =  |\Delta|^2\alpha$, $\hat{\beta} = \beta$ and $\hat{\gamma} =\gamma$
			where ${\bf x_s }= (x_1,x_2)$ denote the horizontal and vertical axis indices and $\Delta$ is the determinant of A.
			
		\end{proposition}
		
		{\bf Proof:} Consider the frequency domain response $\hat{h}$ for view $V$, given by the following.
		\begin{equation}\label{eq:filternew}
		\hat{h}({\bf u}) = \frac{1}{(\alpha \hat{C}({\bf u}) + \beta \hat{D_x}({\bf u}) + \gamma \hat{S_x}({\bf u}))} \hat{M_x}({\bf u}).
		\end{equation}
		For the sake of convenience, we let ${\bf u} = ({\bf u_s},u_3)$ where ${\bf u_s} = (u_1,u_2)$ denotes the spatial frequencies and $u_3$, the temporal frequency. Now using properties of the Fourier transform \cite{bracewell1993affine}, we have,
		
		\begin{eqnarray*}\label{eq:Mx1}
			\hat{M_x}({\bf u_{s}},u_{3})  = \frac{1}{N_e} \sum_{i=1}^{N_e}\hat{X_{i}}({\bf u_{s},u_{3}})\\
			= \frac{1}{N_e} \sum_{i=1}^{N_e} \frac{e^{j2\pi{\bf b} \cdot (A^{-1})^{T}{\bf u_s}} X_{i}((A^{-1})^{T}{\bf u_{s}},u_{3})}{|\Delta|}.\\
		\end{eqnarray*}
		Using the relation $
		M_{x}({\bf u}) = \frac{1}{N_e} \sum_{i=1}^{N_e}X_{i}({\bf u})
		$, we get,
		
		\begin{equation}
		\hat{M_x}({\bf u_{s}},u_{3})  =   \frac{e^{j2\pi{\bf b} \cdot (A^{-1})^{T}{\bf u_s}}M_x((A^{-1})^{T}{\bf u_{s}},u_{3})}{|\Delta|} .
		\end{equation}
		Now,
		
		\begin{eqnarray*}
			\hat{D_x}({\bf u_{s}},u_{3}) = \frac{1}{N_e}\sum_{i=1}^{N_e} |\hat{X_{i}}({\bf u_s},u_3)|^2 \nonumber \\
			= \frac{1}{N_e}\sum_{i=1}^{N_e} |\frac{e^{j2\pi{\bf b} \cdot (A^{-1})^{T}{\bf u_s}}X_i((A^{-1})^{T}{\bf u_s},u_3)}{|\Delta|}|^2 \nonumber \\
			= \frac{1}{N_e}\sum_{i=1}^{N_e} |\frac{X_i((A^{-1})^{T}{\bf u_{s}},u_{3})}{|\Delta|}|^2 .\nonumber
			\vspace{-0.5cm}
		\end{eqnarray*}
		
		Hence, using the relation $
		D_x({\bf u}) = \frac{1}{N_e}\sum_{i=1}^{N_e} |X_{i}({\bf u})|^2
		$, we have
		
		\begin{equation}\label{eq:Dx1}
		\hat{D_x}({\bf u_{s}},u_{3}) = \frac{1}{|\Delta|^2}D_{x}((A^{-1})^{T}{\bf u_{s}},u_{3}).
		\end{equation}
		Similarly, it can be shown that
		
		\begin{equation}\label{eq:Sx1}
		\hat{S_x}({\bf u_{s}},u_{3}) = \frac{1}{|\Delta|^2}S_{x}((A^{-1})^{T}{\bf u_{s}},u_{3}).
		\end{equation}
		Using \eqref{eq:Mx1}, \eqref{eq:Dx1} and \eqref{eq:Sx1} in \eqref{eq:filternew}, we have,
		\begin{small}
			\begin{eqnarray}\label{eq:filternew1}
			\hat{h}({\bf u}) = (e^{j2\pi{\bf b} \cdot (A^{-1})^{T}{\bf u_s}}M_x((A^{-1})^{T}{\bf u_{s}},u_{3}))\Delta \nonumber \\ \frac{1}{(\hat{\alpha}|\Delta|^2 \hat{C}({\bf u}) + \hat{\beta}D_{x}((A^{-1})^{T}{\bf u_{s}},u_{3}) + \hat{\gamma}S_{x}((A^{-1})^{T}{\bf u_{s}},u_{3})}.
			\end{eqnarray}
		\end{small}
		Now letting, $\alpha=\hat{\alpha}|\Delta|^2$, $\beta=\hat{\beta}$, $\gamma=\hat{\gamma}$, $\hat{C}({\bf u})=C({\bf u})=C((A^{-1})^{T}{\bf u_{s}},u_{3}))$ (since $C$ is usually assumed to be equal to 1 at all frequencies if noise model is not available) and using \eqref{eq:filter}, we have,
		
		\begin{equation}
		\hat{h}({\bf u}) = \Delta h((A^{-1})^{T}{\bf u_s},u_3))e^{j2\pi{\bf b} \cdot (A^{-1})^{T}{\bf u_s}}.
		\end{equation}
		Now taking the inverse 3D-FFT of $\hat{h}({\bf u})$, we have, 
		
		\begin{equation}
		\hat{H}({\bf x_{s}},t) = {|\Delta|^2}H(A{\bf x_{s}}+{\bf b},t).
		\label{eq:Affine}
		\end{equation}
		Thus, a compensated MACH filter for the view $V$ is given by ${\bf \hat{H}} =\{\hat{H},\hat{\alpha},\hat{\beta},\hat{\gamma}\}$. This completes the proof of the proposition. Thus a MACH filter for view $V$, with parameters $|\Delta|^2\alpha$, $\beta$ and $\gamma$ can be obtained just by affine transforming the frames of the MACH filter for the canonical view. Normally $|\Delta| \approx 1$ for small view changes. Thus, even though in theory, $\hat{\alpha}$ is related to $\alpha$ by a scaling factor of $|\Delta|^2$, for small view changes, $\hat{h}$ is the optimal filter with essentially the same parameters as those for the canonical view. This result shows that for small view changes, it is possible to build robust MACH filters from a single canonical MACH filter. 
		
\begin{figure}
			\centering
			\includegraphics[width=0.43\textwidth,height=16em]{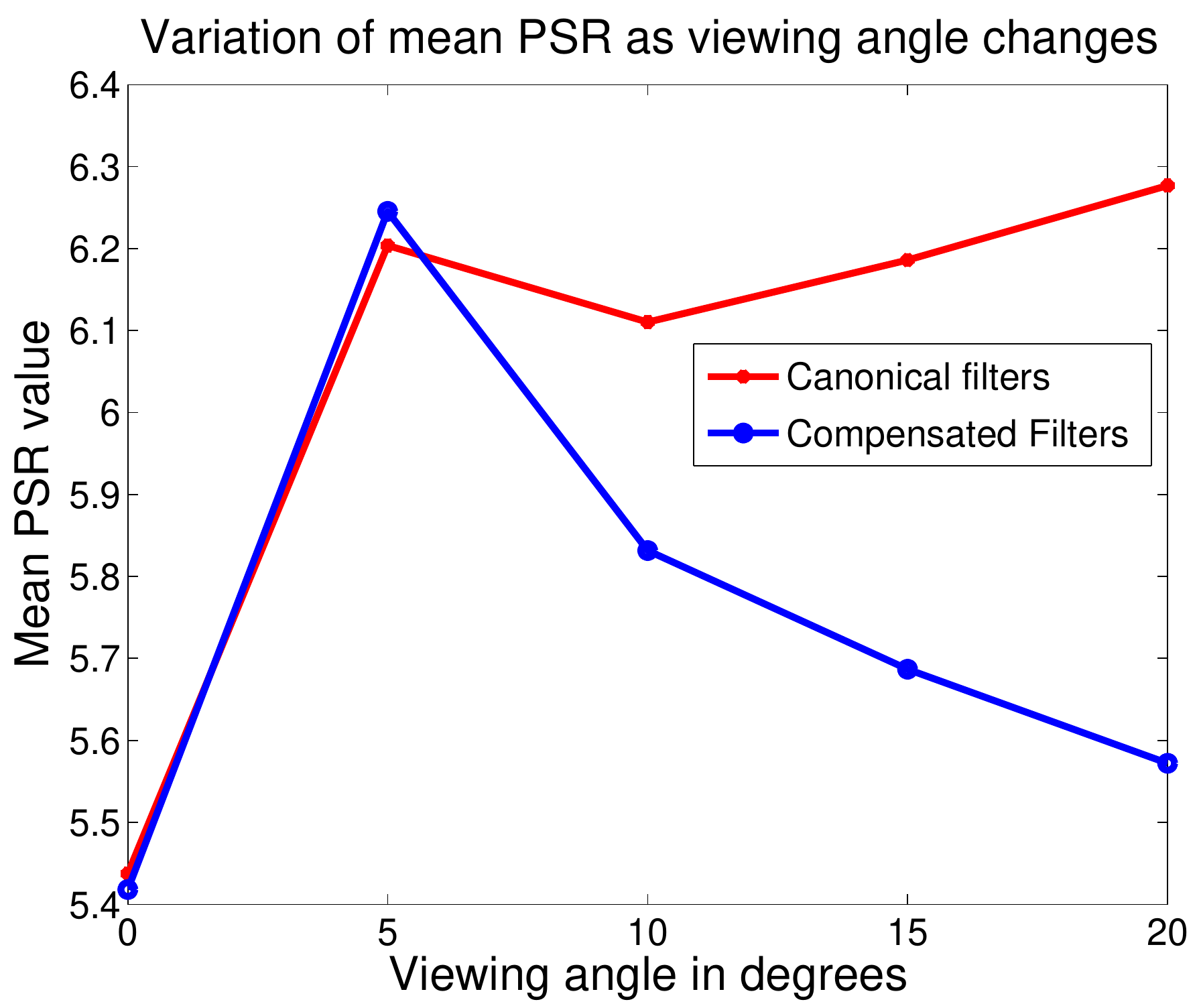}
			\caption{\small{The mean PSRs for different viewpoints for both canonical filters and compensated filters are shown. The mean PSR values obtained using compensated filters are more than those obtained using canonical filters, thus corroborating the need of affine transforming the MACH filters to the viewpoint of the test example.}}
			\label{fig:Affine}
		\end{figure}

		\paragraph*{{\bf Robustness of affine invariant smashed filtering}}
		To corroborate the need of affine transforming the MACH filters to the viewpoint of the test example, we conduct the following two synthetic experiments. In the first, we took all examples in Weizmann dataset and assumed that they belong to the same view, dubbed as the canonical view. We generated five different datasets, each corresponding to a different viewing angle. The different viewing angles from $0^{\circ}$ to $20^{\circ}$ in increments of $5^{\circ}$ were simulated by means of homography. For each of these five datasets, a recognition experiment is conducted using filters for the canonical view as well as the compensated filters for their respective viewpoints, obtained using \eqref{eq:Affine}. The average PSR in both cases for each viewpoint is shown in Figure \ref{fig:Affine}. The mean PSR values obtained using compensated filters are more than those obtained using canonical filters. 
		\par
		In the second experiment, we conducted five independent recognition experiments for the dataset corresponding to fixed viewing angle of $15^{\circ}$, using compensated filters generated for five different viewing angles. The results are tabulated in table \ref{tb:Affine}. It is evident that action recognition rate is highest when the compensated filters used correspond to the viewing angle of the test videos.
		These two synthetic experiments clearly suggest that it is essential to affine transform the filters to the viewpoint of the test video before performing action recognition. 
		\begin{table*}[ht!]
			\centering
			{
				\resizebox{12cm}{!}
				{\begin{tabular}{|l||r|r|r|r|l|}  
						\hline
						Viewing angle & Canonical & $5^{\circ}$ & $10^{\circ}$ & $15^{\circ}$ & $20^{\circ}$ \\
						\hline\hline
						Recognition rate & 65.56 & 68.88 & 67.77 & {\bf 72.22} & 66.67 \\  
						\hline
					\end{tabular}
				}
			}
			\caption{\small{Action recognition rates for the dataset corresponding to fixed viewing angle of $15^{\circ}$ using compensated filters generated for various viewing angles. As expected, action recognition rate is highest when the compensated filters used correspond to the viewing angle of the test videos. }}
			\label{tb:Affine}
		\end{table*} 
		
		\section{Experimental results}
		For all our experiments, we use a measurement matrix, $\phi$ whose entries are drawn from i.i.d. standard Gaussian distribution, to compress the frames of the
		test videos. We conducted extensive experiments on the widely used Weizmann \cite{Weizmann}, UCF sports \cite{MACH}, UCF50 \cite{reddy2013recognizing} and HMDB51 \cite{kuehne2011hmdb} datasets to validate the feasibility of action recognition from compressive cameras. Before we present the action recognition results, we briefly discuss the baseline methods to which we compare our method, and describe a simple to perform action localization in those videos in which the action is recognized
		successfully.
		
		\paragraph*{{\bf Baselines}}
		As noted earlier, this is the first paper to tackle the problem of action recognition from compressive cameras. The absence of precedent approach
		to this problem makes it difficult to decide on the baseline methods to compare with. The state-of-the-art methods for action recognition from traditional
		cameras rely on dense trajectories \cite{wang2013action}, derived using highly non-linear features, HOG \cite{dalal2005histograms}, HOOF \cite{chaudhry2009histograms}, and MBH \cite{wang2011action}. At the moment, it is not quite clear on how to extract such features directly from compressed measurements. Due to these difficulties, we fixate on two baselines. The first baseline method is the Oracle MACH, wherein action recognition is performed as in \cite{MACH} and for the second baseline, we first reconstruct the frames from the compressive measurements using CoSaMP algorithm \cite{needell2010cosamp}, and then apply the improved dense trajectories (IDT) method \cite{wang2013action}, which is the most stable state-of-the-art method, on the reconstructed video to perform action recognition. We use the code made publicly available by the authors, and set all the parameters to default to obtain improved dense trajectory features. The features thus obtained are encoded using Fisher vectors, and a linear SVM is used for classification. Henceforth, we refer this method as Recon+IDT.
		\paragraph*{{\bf Spatial Localization of action from compressive
				cameras without reconstruction}} Action localization in each frame is determined by a bounding box centred at location ($l^{max}$) in that frame, where lmax
		is determined by the peak response (response corresponding to the classified action) in that frame and the size of the filter corresponding to the classified
		action. To determine the size of the bounding box for a particular frame, the response values inside a large rectangle of the size of the filter, and centred
		at $l^{max}$ in that frame are normalized so that they sum up to unity. Treating this normalized rectangle as a 2D probability density function, we determine
		the bounding box to be the largest rectangle centred at $l^{max}$, whose sum is less than a value, $\lambda \le$ 1. For our experiments, we use $\lambda$ equal to 0.7.
		\paragraph*{{\bf Computational complexity}}
		In order to show the substantial computational savings achievable in our STSF framework of reconstruction-free action recognition from compressive cameras, we compare the computational time of the framework with that of Recon+IDT. All experiments are conducted on a
		Intel i7 quad core machine with 16GB RAM.
		\paragraph*{{\bf Compensated Filters}}
		In section 3, we experimentally showed that better action recognition results can be obtained if compensated filters are used instead of canonical view filters (table \ref{tb:Affine}). However, to generate compensated filters, one requires the information regarding the viewpoint of the test video. Generally, the viewpoint of the test video is not known. This difficulty can be overcome by generating compensated filters corresponding to various viewpoints. In our experiments, we restrict our filters to two viewpoints described in section 3, i.e we use `Type-1' and `Type-2' filters.
		
		\subsection{Reconstruction-free recognition on Weizmann dataset}
		Even though it is widely accepted in the computer vision community that Weizmann dataset is an easy dataset, with many methods achieving near perfect
		action recognition rates, we believe that working with compressed measurements precludes the use of those well-established methods, and obtaining such high
		action recognition rates at compression ratios of 100 and above even for a simple dataset as Weizmann is not straightforward. The Weizmann dataset contains
		10 different actions, each performed by 9 subjects, thus making a total of 90 videos. For evaluation, we used the leave-one-out approach, where the filters
		were trained using actions performed by 8 actors and tested on the remaining one. The results shown in table \ref{tb:Weiz1} indicate that our method clearly outperforms
		the Recon+IDT. It is quite evident that with full-blown frames (indicated in table \ref{tb:Weiz1}) that Recon+IDT method performs much better than STSF method. However,
		at compression ratios of 100 and above, recognition rates are very stable for our STSF framework, while Recon+IDT fails completely. This is due to the fact
		that Recon+IDT operates on reconstructed frames, which are of poor quality at such high compression ratios, while STSF operates directly on compressed
		measurements. The recognition rates are stable even at high compression ratios and are comparable to the recognition accuracy for the Oracle MACH (OM) method \cite{MACH1}.
		\begin{table}[ht!]
			\flushleft
			{
				\resizebox{8.4cm}{1.1cm} {
					\begin{tabular}{|c|c|c|p{2cm}|}  
						\hline
						Compression factor & STSF & Recon + IDT\\
						\hline\hline
						1  & 81.11 (3.22s) (OM \cite{MACH1,MACH} ) & 100 (3.1s) \\
						\hline
						100 & 81.11 (3.22s)  & 5.56 (1520s)\\
						\hline
						200 & 81.11 (3.07s) & 10 (1700s)\\
						\hline
						300 & 76.66 (3.1s) & 10 (1800s) \\
						\hline
						500 & 78.89 (3.08s) & 7.77 (2000s) \\
						\hline
					\end{tabular}
				}
			}
			\caption{\small{Weizmann dataset: Recognition rates for reconstruction-free
					recognition from compressive cameras for different compression factors are stable even at high compression factors of 500. Our method clearly outperforms Recon+IDT
					method and is comparable to Oracle MACH \cite{MACH1,MACH}.}}
			\label{tb:Weiz1}
		\end{table}
		
\begin{figure*}[ht!]
			\centering
			\subfigure[]{\includegraphics[width=0.9\textwidth, height= 0.6in]{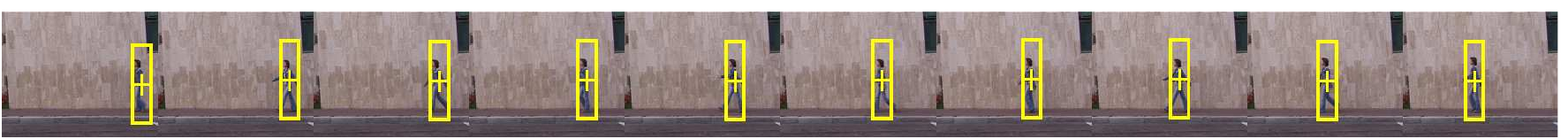}
				\label{walkloc}
			} 
			\subfigure[]{\includegraphics[width=0.9\textwidth, height = 0.6in]{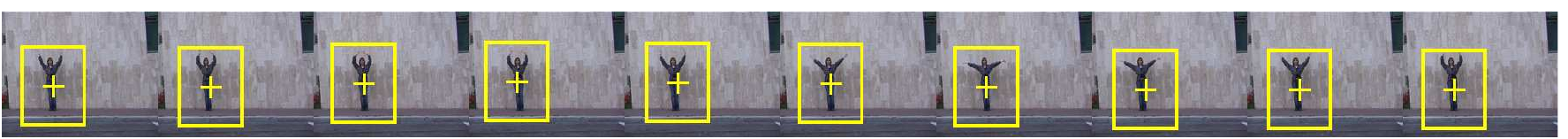}
				\label{wave1loc}
			}
			\subfigure[]{\includegraphics[width=0.9\textwidth,height = 0.6in]{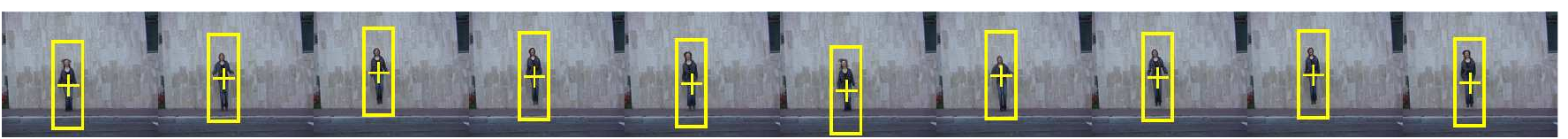}
				\label{pjumploc}
			}
			\caption{\small{Spatial localization of subject without reconstruction at compression ratio = 100 for different actions in Weizmann dataset. a) Walking b) Two handed wave c) Jump in place}}
			\label{fig:Actloc}
		\end{figure*}

		The average time taken by STSF and Recon+IDT to process a video of size $144 \times 180 \times 50$ are shown in parentheses in table 1. Recon+IDT takes about 20-35 minutes to process one video, with the frame-wise reconstruction of the video being the dominating component in the total computational time, while STSF framework takes only a few seconds for the same sized video since it operates directly on compressed measurements.
		\paragraph*{{\bf Spatial localization of action from compressive cameras without reconstruction}}
		Further, to validate the robustness of action detection using the STSF framework, we quantified action localization in terms of error in estimation of the subject's centre from its ground truth. The subject's centre in each frame is estimated as the centre of the fixed sized bounding box with location of the peak response (only the response corresponding to the classified action) in that frame as it left-top corner. Figure \ref{fig:Actloc} shows action localization in a few frames for various actions of the dataset (More action localization results for Weizmann dataset can be found in supplementary material). Figure \ref{fig:Actlocerr} shows that using these raw estimates, on average, the error from the ground truth is less than or equal to 15 pixels in approximately 70\% of the frames, for compression ratios of 100, 200 and 300. It is worth noting that  using our framework it is possible to obtain robust action localization results without reconstructing the images, even at extremely high compression ratios.
	\paragraph*{Experiments on the {\bf KTH} dataset} 
	We also conducted experiments on the KTH dataset \cite{KTH}. Since the KTH dataset is considered somewhat similar to the Weizmann dataset in terms of the difficulty and scale, we have relegated the action recognition results to the supplement.
		
		\begin{figure*}[ht!]
			\centering
			\subfigure[]{\includegraphics[height=8em,width=0.215\textwidth]{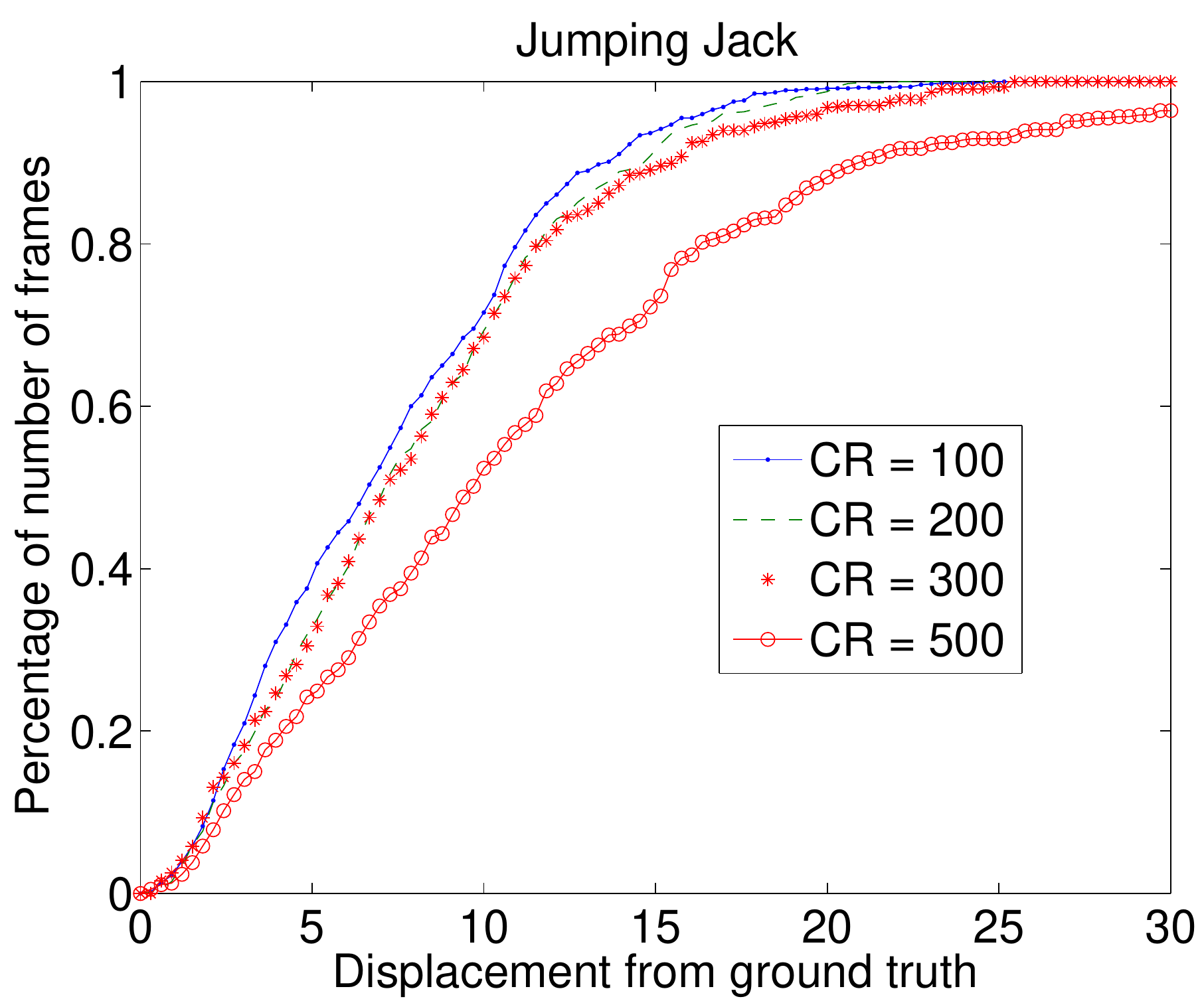}
				\label{fig:subfig1}
			}
			\subfigure[]{\includegraphics[height=8em,width=0.215\textwidth]{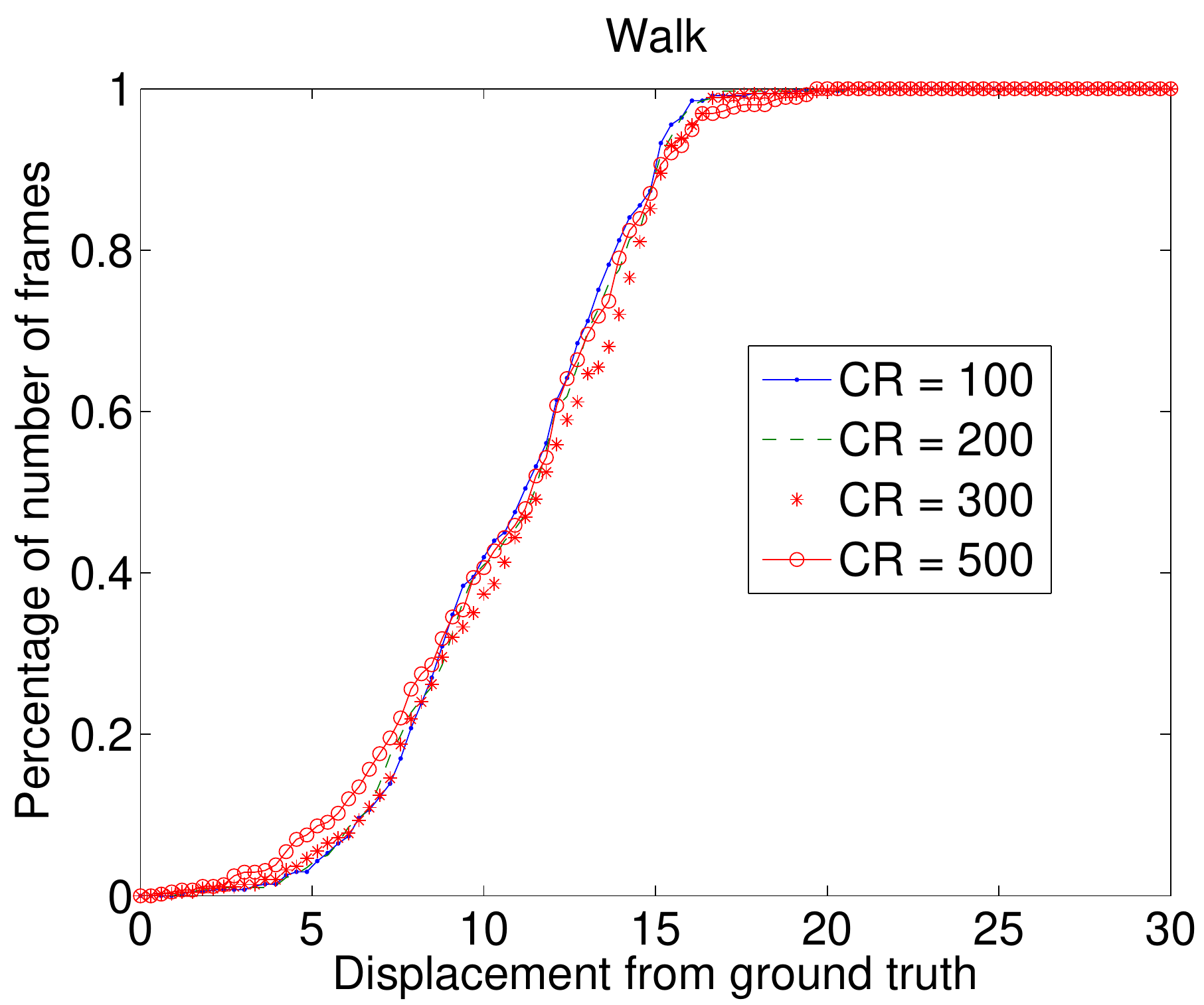}
				\label{fig:subfig2}
			}
			\subfigure[]{\includegraphics[height=8em,width=0.215\textwidth]{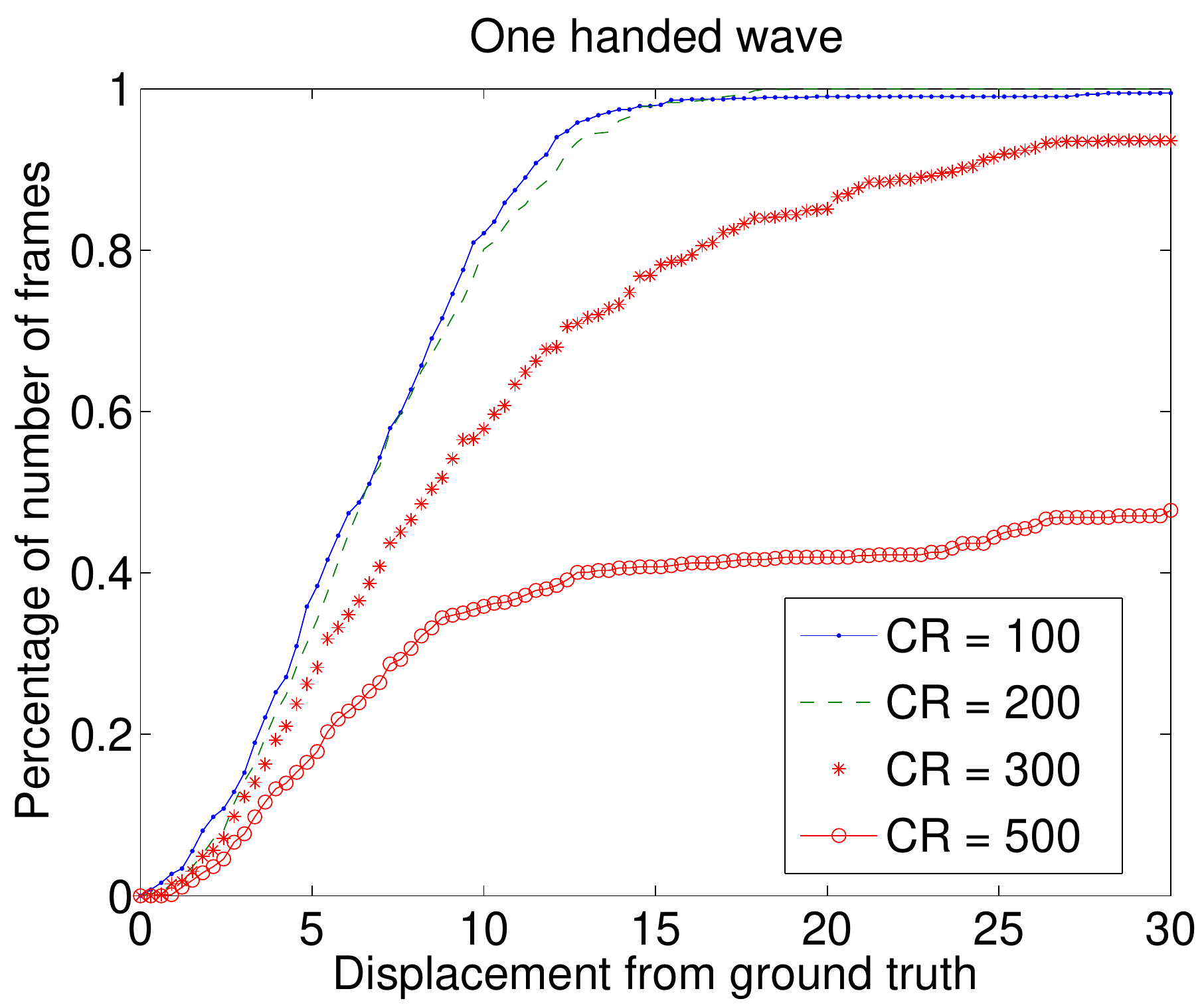}
				\label{fig:subfig1}
			}
			\subfigure[]{\includegraphics[height=8em,width=0.215\textwidth]{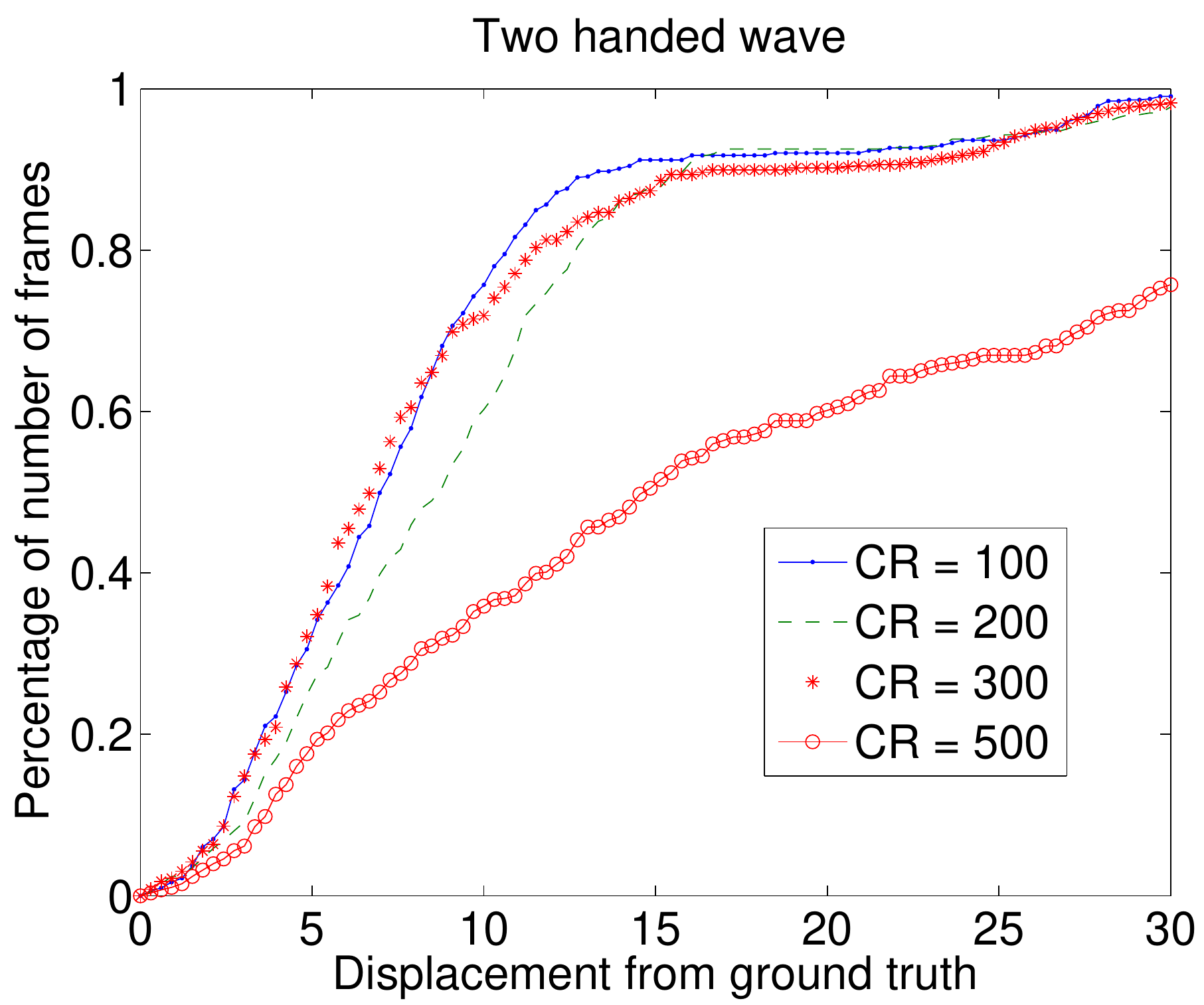}
				\label{fig:subfig1}
			}
			\caption{\small{Localization error for Weizmann dataset. X-axis : Displacement from ground truth. Y-axis: Fraction of total number of frames for which the displacement of subject's centre from ground truth is less than or equal to the value in x-axis. On average, for approximately 70\% of the frames, the displacement of ground truth is less than or equal to 15 pixels, for compression ratios of 100, 200 and 300.}}
			\label{fig:Actlocerr}
			
		\end{figure*}

		\subsection{Reconstruction-free recognition on UCF sports dataset}
		The UCF sports action dataset \cite{MACH} contains a total of 150 videos across 9 different actions. The dataset is a challenging dataset with scale and viewpoint variations. For testing, we use leave-one-out cross validation. At compression ratio of 100 and 300, the recognition rates are 70.67\% and 68\% respectively. The rates obtained are comparable to those obtained in Oracle MACH set-up \cite{MACH} (69.2\%). Considering the difficulty of the dataset, these results are very encouraging. The confusion matrices for compression ratios 100 and 300 are shown in tables \ref{tb:UCF100} and \ref{tb:UCF300} respectively.
		\begin{table*}[ht!]
			\centering
			{
				\resizebox{16.6cm}{2.1cm}
				{\begin{tabular}{|l||r|r|r|r|r|r|r|r|l|}  
						\hline
						Action & Golf-Swing & Kicking & Riding Horse & Run-Side & Skate-Boarding  & Swing & Walk & Diving & Lifting\\
						\hline\hline
						Golf-Swing & {\bf 77.78} & 16.67 & 0 & 0 & 0 & 0 & 5.56 & 0 & 0 \\  
						\hline
						Kicking & 0 & {\bf 75} & 0  & 5 & 5 & 10 & 5 & 0 & 0\\  
						\hline
						Riding Horse & 16.67 & 16.67 & {\bf 41.67} & 8.33 & 8.33 & 0 & 8.33 & 0 & 0\\  
						\hline
						Run-Side & 0 & 0 & 0 & {\bf 61.54} & 7.69 & 15.38 & 7.69 & 7.69 & 0 \\  
						\hline
						Skate-Boarding & 0 & 8.33 & 8.33 & 25 & {\bf 50} & 0 & 5 & 0 & 0\\  
						\hline
						Swing & 0 & 3.03 & 12.12 & 0.08 & 3.03 & {\bf 78.79} & 3.03 & 0 & 0\\  
						\hline
						Walk & 0 & 9.09 & 4.55 & 4.55 & 9.09 & 9.09 & {\bf 63.63} & 0 & 0\\  
						\hline
						Diving & 0 & 0 & 0 & 0 & 7.14 & 0 & 0 &  {\bf 92.86} & 0\\  
						\hline
						Lifting & 0 & 0 & 0 & 0 & 0 & 0 & 0 &  16.67 & {\bf 83.33}\\  
						\hline
					\end{tabular}
				}
			}
			\caption{\small{Confusion matrix for UCF sports database at a compression factor = 100. Recognition rate for this scenario is 70.67 \%, which is comparable to Oracle MACH \cite{MACH} (69.2\%).}}
			\label{tb:UCF100}
		\end{table*}
		\begin{table*}[ht!]
			\centering
			{
				\resizebox{16.6cm}{2.1cm}
				{\begin{tabular}{|l||r|r|r|r|r|r|r|r|l|}  
						\hline
						Action & Golf-Swing & Kicking & Riding Horse & Run-Side & Skate-Boarding  & Swing & Walk & Diving & Lifting\\
						\hline\hline
						Golf-Swing & {\bf 55.56} & 0 & 27.78 & 0 & 0 & 5.56 & 11.11 & 0 & 0 \\  
						\hline
						Kicking & 0 & {\bf 95} & 0  & 5 & 0 & 0 & 0 & 0 & 0\\  
						\hline
						Riding Horse & 0 & 0 & {\bf 75} & 16.67 & 0 & 8.33 & 0 & 0 & 0\\  
						\hline
						Run-Side & 0 & 0 & 7.69 & {\bf 38.46} & 7.69 & 30.77 & 7.69 & 7.69 & 0 \\  
						\hline
						Skate-Boarding & 8.33 & 0 & 0 & 8.33 & {\bf 50} & 16.67 & 16.67 & 0 & 0\\  
						\hline
						Swing & 0 & 0 & 0 & 12.12 & 12.12 & {\bf 72.73} & 3.03 & 0 & 0\\  
						\hline
						Walk & 0 & 0 & 0 & 0 & 4.55 & 22.73 & {\bf 72.73} & 0 & 0\\  
						\hline
						Diving & 0 & 0 & 14.29 & 14.29 & 0 & 14.29 & 0 &  {\bf 57.14} & 0\\  
						\hline
						Lifting & 0 & 0 & 0 & 0 & 0 & 0 & 0 &  16.67 & {\bf 83.33}\\  
						\hline
					\end{tabular}
				}
			}
			\caption{\small{Confusion matrix for UCF sports database at a compression factor = 300. Recognition rate for this scenario is 68 \%.}}
			\label{tb:UCF300}
		\end{table*}
		
\begin{figure*}[ht!]
			\centering
			\vspace{-0.5cm}
			\subfigure[]{\includegraphics[width=1\textwidth, height= 1.2in]{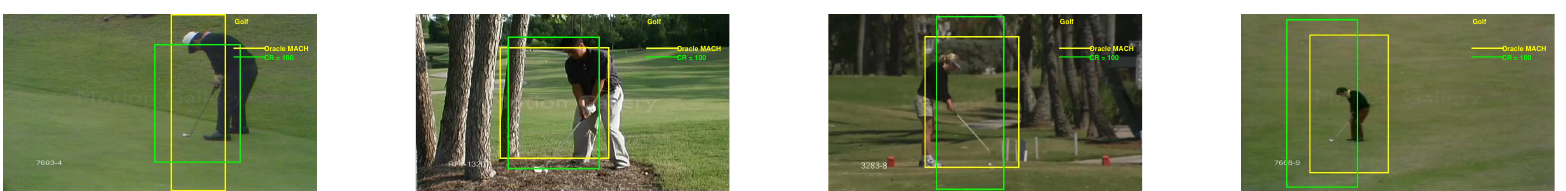}
				\label{golfloc}
			} 
			\vspace{-0.4cm}
			\subfigure[]{\includegraphics[width=1\textwidth, height = 1.2in]{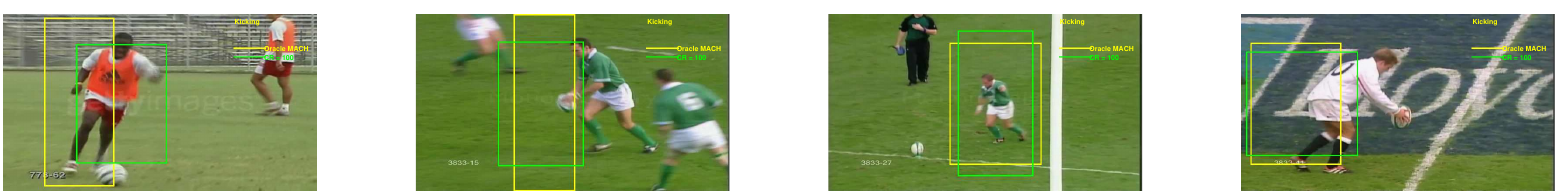}
				\label{kickloc}
			}
			\vspace{-0.4cm}
			\subfigure[]{\includegraphics[width=1\textwidth,height = 1.2in]{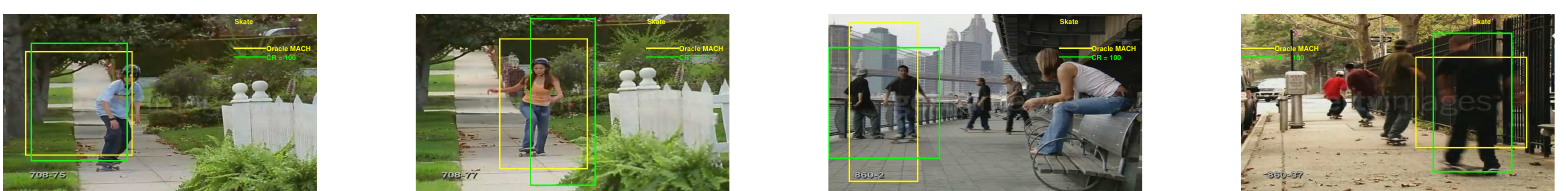}
				\label{skateloc}
			}
			\vspace{-0.2cm}
			\caption{\small{Reconstruction-free spatial localization of subject for Oracle MACH (shown as yellow box) and STSF (shown as green box) at compression ratio = 100 for some correctly classfied instances of various actions in the UCF sports dataset. a) Golf b) Kicking c) Skate-Boarding. Action localization is estimated reasonably well directly from CS measurements even though the measurements themselves do not bear any explicit information regarding pixel locations. }}
			\label{fig:ActlocUCF}
		\end{figure*}

		\paragraph*{{\bf Spatial localization of action from compressive cameras without reconstruction}}
		Figure \ref{fig:ActlocUCF} shows action localization for some correctly classified instances across various actions in the dataset, for Oracle MACH and compression ratio = 100 (More action localization results can be found in supplementary material). It can be seen that action localization is estimated reasonably well despite large scale variations and extremely high compression ratio. 
		
		\subsection{Reconstruction-free recognition on UCF50 dataset}
\begin{figure*}
			\centering
			\includegraphics[width=\textwidth,height=62em]{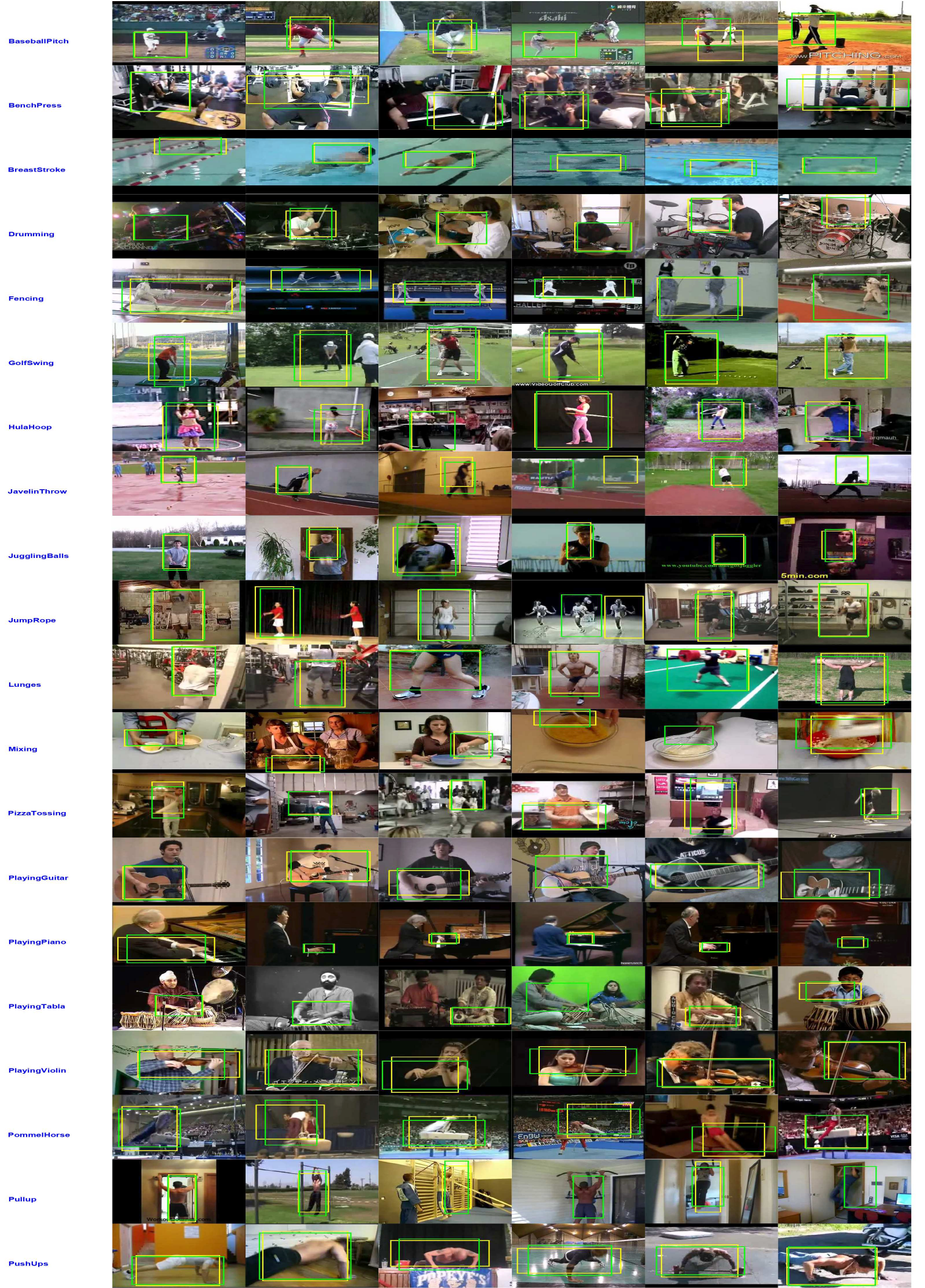}
			\caption{\small{Action localization: Each row corresponds to various instances of a particular action, and action localization in one frame for each of these instances is shown. The bounding boxes (yellow for Oracle MACH, and green for STSF at compression ratio = 100) in most cases correspond to the human, or the moving part. Note that these bounding boxes shown are obtained using a rudimentary procedure, without any training, as outlined earlier in the section. This suggests that joint training of features extracted from correlation volumes and annotated bounding boxes can lead to more accurate action localization results.}}
			\label{fig:Act_UCF50}
		\end{figure*}

		To test the scalability of our approach, we conduct action recognition on large datasets, UCF50 \cite{reddy2013recognizing} and HMDB51 \cite{kuehne2011hmdb}. Unlike the datasets considered earlier, these two datasets have large intra-class scale variability. To account for this scale variability, we generate about 2-6 filters per action. To generate MACH filters, one requires bounding box annotations for the videos in the datasets. Unfortunately frame-wise bounding box annotations are not available for these two datasets. Hence, we manually annotated a large number of videos in the UCF50 dataset. In total, we generated 190 filters, as well as their flipped versions
		(`Type-2' filters). The UCF50 database consists of 50 actions, with around 120 clips per action, totaling upto 6681 videos. The database is divided into 25 groups with
		each group containing between 4-7 clips per action. We use leave-one-group cross-validation to evaluate our framework. The recognition rates at different compression ratios, and the mean time taken for one clip (in
		parentheses) for our framework and Recon+IDT are tabulated in table \ref{tb:UCF50}. Table \ref{tb:UCF50} also shows the recognition rates for various state-of-the-art action recognition methods, while operating on the full-blown images, as indicated in the table by (FBI).  Two conclusions follow from the table. 1) Our approach outperforms the baseline method, Recon+IDT at very high compression ratios of 100 and above, and 2) the mean time per clip is less than that for Recon+IDT method. This clearly suggests that when operating at high compression ratios, it is better to perform action recognition without reconstruction than reconstructing the frames and then applying a state-of-the-art method.   The recognition rates for individual classes for Oracle MACH (OM), and compression ratios, 100 and 400 are given in table \ref{tb1:UCF50}. The action localization results for various actions
		are shown in figure \ref{fig:Act_UCF50}. The bounding boxes in most instances correspond to the human or the moving part of the human or the object of interest. Note
		how the sizes of the bounding boxes are commensurate with the ‘area’ of the action in each frame. For example, for the fencing action, the bounding box
		covers both the participants, and for the playing piano action, the bounding box covers just the hand of the participant. In the case of breaststroke action, where human is barely visible, action localization results are impressive. We emphasize that action localization is achieved directly from compressive measurements without any intermediate reconstruction, even though the measurements do not bear any explicit information regarding pixel locations. We note that the procedure outlined above is by no means a full-fledged procedure for action localization and is  fundamentally different from the those in \cite{lan2011discriminative,tianspatiotemporal}, where sophisticated models are trained jointly on action labels and the location of person in each frame, and action and its localization are determined simultaneously by solving one computationally intensive inference problem. While our method is simplistic in nature and does not always estimate localization accurately, it relies only on minimal post-processing of the correlation response, which makes it an attractive solution for action localization in resource-constrained environments where a rough estimate of action location may serve the purpose. However, we do note that action localization is not the primary goal of the paper and that the purpose of this exercise is to show that reasonable localization results directly from compressive measurements are possible, even using a rudimentary procedure as outlined above. This clearly suggests that with more sophisticated models, better reconstruction-free action localization results can be achieved. One possible option is to co-train models jointly on action labels and annotated bounding boxes in each frame similar to \cite{lan2011discriminative,tianspatiotemporal}, while extracting spatiotemporal features such as HOG3D \cite{klaser2008spatio} features for correlation response volumes, instead of the input video. 
		
		\begin{table}[ht!]
			\centering
			{
				\resizebox{8.3cm}{1.1cm}
				{
					\begin{tabular}{|c|c|c|c|}  
						\hline
						Method & CR = 1 & CR = 100 & CR =400\\
						\hline
						Our method (`Type 1' + `Type 2') & 60.86 (2300s) (OM) & 54.55 (2250s) & 46.48 (2300s) \\
						\hline
						Recon + IDT & 91.2 (FBI) & 21.72 (3600s) & 12.52 (4000s) \\
						\hline
						Action Bank \cite{AcBank2012}  & 57.9 (FBI) & NA & NA \\
						\hline
						Jain \etal \cite{jain2013representing} & 59.81 (FBI) & NA & NA \\
						\hline
						Kliper-Gross \etal \cite{kliper2012motion} & 72.7 (FBI) & NA & NA \\
						\hline 
						Reddy \etal \cite{reddy2013recognizing} & 76.9 (FBI) & NA & NA \\
						\hline
						Shi \etal \cite{shi2013sampling} & 83.3 (FBI) & NA & NA \\
						\hline
					\end{tabular}
				}
			}
			\caption{\small{UCF50 dataset: The recognition rate for our framework is stable even at very high compression ratios, while in the case of Recon + IDT, it falls off spectacularly. The mean time per clip (given in parentheses) for our method is less than that for the baseline method (Recon + IDT).}}
			\label{tb:UCF50}
		\end{table}
		
		\begin{table*}[ht!]
			\centering
			{
				\resizebox{16.6cm}{1.6cm}
				{\begin{tabular}{|l||r|r|r||r|r|r|r||r|r|r|r||r|r|r|r|}  
						\hline
						Action & CR =1 (OM) & CR = 100 & CR = 400 & Action & CR =1 (OM) & CR = 100 & CR = 400 & Action & CR =1 (OM) & CR = 100 & CR = 400 & Action & CR =1 (OM) & CR = 100 & CR = 400\\
						\hline
						BaseballPitch & 58.67 & 57.05 & 50.335 & HorseRiding & 77.16 & 60.4 & 60.4 & PlayingPiano & 65.71 & 60.95 & 58.1 & Skiing & 35.42 & 34.72 & 29.86\\  
						\hline
						Basketball & 41.61 & 38.2353 & 25.7353 & HulaLoop & 55.2 & 56 & 55.2 & PlayingTabla & 73.88 & 56.75 & 36.94 & Skijet & 44 & 37 & 29\\  
						\hline
						BenchPress & 80 & 73.75 & 65.63 & Javelin Throw & 41.0256 & 41.0256 & 32.48 & PlayingViolin & 59 & 52 & 43 & SoccerJuggling & 42.31 & 31.61 & 28.38\\  
						\hline
						Biking & 60 & 42.07 & 33.01 & Juggling Balls & 64.75 & 67.21 & 65.57 & PoleVault & 56.25 & 58.12 & 53.75 & Swing & 54.01 & 35.03 & 19.7 \\  
						\hline
						Billiards & 94.67 & 89.33 & 79.33 & JumpRope & 71.53 & 75 & 74.31 & PommelHorse & 86.07 & 81.3 & 69.1 & TaiChi & 66 & 68 & 61\\  
						\hline
						Breaststroke & 81.19 & 46.53 & 17.82 & JumpingJack & 80.49 & 80.49 & 72.357 & PullUp & 64 & 59 & 49 & TennisSwing & 46.11 & 41.92 & 30.53 \\  
						\hline
						CleanAndJerk & 56.25 & 59.82 & 41.96 & Kayaking & 58.6 & 47.14 & 43.12 & Punch & 80.63 & 73.12 & 62.5 & ThrowDiscus & 62.6 & 51.14 & 45\\  
						\hline
						Diving & 76.47 & 71.24 & 51.63 & Lunges & 44.68 & 36.17 & 32.62 &  PushUps & 66.67 & 60.78 & 61.76 & TrampolineJumping & 45.39 & 28.57 & 18.48\\  
						\hline
						Drumming & 63.35 & 50.93 & 44.1 & MilitaryParade & 80.32 & 78.74 & 59.05 &  RockClimbing & 65.28 & 58.33 & 63.2 & VolleyBall & 60.34 & 48.27 & 39.65 \\  
						\hline
						Fencing & 71.171 & 64.86 & 62.16 & Mixing & 51.77 & 56.02 & 48.93 &  RopeClimbing & 36.92 & 34.61 & 29.23 & WalkingwithDog & 31.71 & 27.64 & 25.4 \\  
						\hline
						GolfSwing & 71.13 & 58.86 & 48.93 & Nunchucks & 40.9 & 34.1 & 31.82 &  Rowing & 55.47 & 40.14 & 29.2 & YoYo & 54.69 & 58.59 & 47.65\\  
						\hline
						HighJump & 52.03 & 52.84 & 47.15 & Pizza Tossing & 30.7 & 33.33 & 22.8 &  Salsa & 69.92 & 63.16 & 46.62 & & & &\\  
						\hline
						HorseRace & 73.23 & 66.92 & 59.84 & PlayingGuitar & 73.75 & 64.37 & 60.62 &  SkateBoarding & 55.82 & 46.67 & 38.33 & & & &\\  
						\hline
					\end{tabular}
				}
			}
			\caption{\small{UCF50 dataset: Recognition rates for individual classes at compression ratios, 1 (Oracle MACH), 100 and 400.}}
			\label{tb1:UCF50}
		\end{table*}

		\subsection{Reconstruction-free recognition on HMDB51 dataset}
		The HMDB51 database consists of 51 actions, with around 120 clips per action, totalling upto 6766 videos. The database is divided into three
		train-test splits. The average recognition rate across these splits is reported here. For HMDB51 dataset, we use the same filters which were generated for UCF50 dataset. The recognition rates at different compression ratios, and mean time taken
		for one clip (in parentheses) for our framework and Recon+IDT are tabulated in table \ref{tb:HMDB51}. Table \ref{tb:HMDB51} also shows the recognition rates for various state-of-the-art
		action recognition approaches, while operating on full-blown images. The table clearly suggests that while operating at compression ratios of 100 and above, to perform action recognition, it is better to work in compressed domain rather than reconstructing the frames, and then applying a state-of-the-art method.
		\begin{table}[ht!]
			\centering
			{
				\resizebox{8.3cm}{1cm}
				{
					\begin{tabular}{|c|c|c|c|}  
						\hline
						Method & CR = 1 & CR = 100 & CR =400\\
						\hline
						Our method (`Type 1' + `Type 2') & 22.5 (2200s) (OM) & 21.125 (2250s) & 17.02 (2300s) \\
						\hline
						Recon + IDT & 57.2 (FBI) & 6.23 (3500s) & 2.33 (4000s) \\
						\hline
						Action Bank \cite{AcBank2012}  & 26.9 (FBI) & NA & NA \\
						\hline
						Jain \etal \cite{jain2013better} & 52.1 (FBI) & NA & NA \\
						\hline
						Kliper-Gross \etal \cite{kliper2012motion} & 29.2 (FBI) & NA & NA \\
						\hline 
						Jiang \etal \cite{jiang2012trajectory} & 40.7 (FBI) & NA & NA \\
						\hline
					\end{tabular}
				}
			}
			\caption{\small{HMDB51 dataset: The recognition rate for our framework is stable even at very high compression ratios, while in the case of Recon+IDT, it falls off spectacularly.}}
			\label{tb:HMDB51}
		\end{table}
		
		\section{Discussions and Conclusion}
		In this paper, we proposed a correlation based framework to recognize actions from compressive cameras without reconstructing the sequences. It is
		worth emphasizing that the goal of the paper is not to outperform a state-of-the-art action recognition system but is to build a action recognition system
		which can perform with an acceptable level of accuracy in heavily resource-constrained environments, both in terms of storage and computation. The fact that we are able to achieve a recognition rate of 54.55\% at a compression ratio of 100 on a difficult and large dataset like UCF50 and also localize the actions
		reasonably well clearly buttresses the applicability and the scalability of reconstruction-free recognition in resource constrained environments. Further, we reiterate that at compression ratios of 100 and above, when reconstruction is generally of low quality, action recognition results using our approach, while working in compressed domain, were shown to be far better than reconstructing the images, and then applying a state-of-the-art method. In our future research, we wish to extend this approach to more generalizable filter-based approaches. One possible extension is to use motion sensitive filters like Gabor or Gaussian
		derivative filters which have proven to be successful in capturing motion. Furthermore, by theoretically proving that a single filter is sufficient to encode an action over the space of all affine transformed views of the action, we showed that more robust filters can be designed by transforming all training examples to a canonical viewpoint.

		
		%


		\ifCLASSOPTIONcompsoc
		\section*{Acknowledgments}
		\else
		\section*{Acknowledgment}
		\fi
		
		The authors would like to thank Henry Braun for their useful suggestions and comments. 
		
		\ifCLASSOPTIONcaptionsoff
		\newpage
		\fi
		
		
		
		
		\bibliographystyle{IEEEtran}
		%
		\bibliography{PAMI_STSF_3}

		
		
		

		%

		\begin{IEEEbiography}{Kuldeep Kulkarni}
			
		\end{IEEEbiography}
		
		\begin{IEEEbiography}{Pavan Turaga}
			
		\end{IEEEbiography}
		
		
		

	\end{document}